\documentclass[10pt,leqno]{amsart}
\usepackage{graphicx}
\usepackage{indentfirst,csquotes}

\usepackage{subfigure}
\usepackage{cite}
\usepackage{amssymb,amsthm,amsmath}
\usepackage{array}
\usepackage{rotating}
\usepackage{graphicx}
\usepackage{amsmath}
\usepackage{multirow}
\usepackage{multicol}
\usepackage{longtable}
\usepackage{siunitx}
\usepackage{booktabs}
\usepackage{adjustbox}
\newcommand{\etal}{\textit{et al.}}

\AtBeginDocument{%
  }
\usepackage{xcolor,paralist,hyperref,titlesec,fancyhdr,etoolbox}

\titleformat{\section}[display]{\normalfont\huge\bfseries\centering}{\centering\chaptertitlename\thechapter}{10pt}{\Large}
\titlespacing*{\section}{0pt}{0ex}{0ex}

\hypersetup{ colorlinks=true, linkcolor=black, filecolor=black, urlcolor=black }

\usepackage{lipsum}

\begin{document}
\title{Green Federated Learning: A new era of Green Aware AI} 
\author{Dipanwita Thakur}

\author{Antonella Guzzo}
\author{Giancarlo Fortino}

\author{Francesco Piccialli}
   
\maketitle

\let\thefootnote\relax
\footnotetext{DIMES - University of Calabria, Italy and Department of Mathematics and Applications "R. Caccioppoli", University of Naples Federico II, Italy} 

\begin{abstract}
The development of AI applications, especially in large-scale wireless networks, is growing exponentially, alongside the size and complexity of the architectures used. Particularly, machine learning is acknowledged as one of today's most energy-intensive computational applications, posing a significant challenge to the environmental sustainability of next-generation intelligent systems. Achieving environmental sustainability entails ensuring that every AI algorithm is designed with sustainability in mind, integrating green considerations from the architectural phase onwards. Recently, Federated Learning (FL), with its distributed nature, presents new opportunities to address this need. Hence, it's imperative to elucidate the potential and challenges stemming from recent FL advancements and their implications for sustainability. Moreover, it's crucial to furnish researchers, stakeholders, and interested parties with a roadmap to navigate and understand existing efforts and gaps in green-aware AI algorithms. This survey primarily aims to achieve this objective by identifying and analyzing over a hundred FL works, assessing their contributions to green-aware artificial intelligence for sustainable environments, with a specific focus on IoT research. It delves into current issues in green federated learning from an energy-efficient standpoint, discussing potential challenges and future prospects for green IoT application research.
\end{abstract} 

\bigskip

\section*{Introduction}
\subsection*{Contest Scenario}
Since 2012, the discipline of artificial intelligence of artificial intelligence (AI) has made impressive strides in various areas, including speech recognition, object recognition, machine translation, and game playing. Machine learning models that are getting bigger and more computationally intensive have contributed significantly to this advancement. Initially, AI research focuses on the accuracy improvement of the model using enormous computational energy, leading to Red AI \cite{Roy2020}. Hence, the computational cost is very high, and due to massive power consumption, Red AI is not environment-friendly, leading to \textbf{``Green AI''} - AI research that is more inclusive and friendly to the environment. The phrase ``Green AI'' relates to AI research that produces new insights while accounting for the computational cost, promoting decreased resource use.
In contrast to Red AI, Green AI encourages methods with favorable performance/efficiency trade-offs, which has led to fast-rising computational (and consequently carbon) costs. Researchers will have the choice to focus on the efficiency of their models with a positive influence on both inclusiveness and the environment if efficiency measurements are widely acknowledged as essential assessment metrics for research along with accuracy. Traditional machine learning approaches are unsuitable for green AI, as traditional ML algorithms need centralized data collection and processing by a centralized server, which is becoming a bottleneck of large-scale applications in daily life due to progressively growing privacy issues. 
\par

``Federated learning" (FL) is a recent distributed learning paradigm where a central server coordinates a global model's cooperative training by several clients. Federated learning keeps the training data local while allowing resource-constrained edge computing devices, like IoT and smartphones, to create a common model for prediction. Clients must submit updated models (such as those represented by gradients or parameters) to the server after training the models on their private local data. Cross-device FL and cross-silo FL are two categories of FL that can be distinguished by the customers who participate and the training scale. Clients in cross-device FL are tiny, dispersed entities (smartphones, wearables, and edge devices, for example), and each client probably has a modest quantity of local data. Therefore, for cross-device FL to succeed, a sizable number of edge devices—up to millions—must participate in the training process. However, in cross-silo FL, clients are usually businesses or institutions (banks, hospitals, etc.). There are very few participants—between two and one hundred, perhaps—and every client is required to be present for the duration of the session \cite{Huang2022crosssilo}. 

Federated Learning (FL) 
may also be resource-intensive and have a high energy consumption, especially when used on a scale. 
In the smart environment, billions of IoT devices, including smartphones, laptops, wearable technology, and cars, are connected by IoT networks \cite{Al-Abiad2021energy}. FL at the edge should concentrate on reducing latency and resource energy usage without compromising the global model's convergence rate in order to boost system reliability and advance green IoT. In terms of energy-efficient FL, reducing the energy consumption of edge devices is crucial to maintaining the long-term sustainability of the FL framework because mobile devices run on batteries. Reducing communication and computing energy should get significant consideration in a green FL or energy-efficient system.
To reduce energy consumption and achieve an optimized model in terms of performance and efficiency, \textbf{Green Federated Learning} is introduced by the research community, which ultimately promotes sustainable AI. Green federated learning addresses two significant issues - (i) Privacy concerns: a central server, and local end devices maintain the same model by exchanging model updates instead of raw data, with which the privacy of data stored on end devices is not directly revealed, and (ii) Huge energy consumption issue: Green FL involves choosing design elements and FL parameters to decrease energy consumption while maintaining competitive performance and training time.

\subsection*{Motivation}
Table \ref{tab:relatedsurvey} summarizes pertinent surveys to demonstrate our motivation.
\begin{table}
\centering
\caption{A Brief Summary of Related Surveys on Federated Learning}
    \label{tab:relatedsurvey} 
    \begin{adjustbox}{width=1\textwidth}
\noindent\begin{tabular}{lSSSSSSSSSSSl}
\toprule
& \multicolumn{10}{c}{Focus} &  \\
\cmidrule(r){2-12}
Survey & \multicolumn{4}{c}{Heterogeneity} &  \multicolumn{6}{c}{Efficiency} & Application & Green AI\\
\cmidrule(r){2-5}\cmidrule(l){6-11}
& \multicolumn{3}{c}{Statistical} & {System} &  \multicolumn{5}{c}{Communication} & {Computation} & &\\
\cmidrule(r){2-4}\cmidrule(l){6-10}
& A1 & A2 & A3 & {} &  B1 & \multicolumn{3}{c}{B2} & B3 & {} & &\\
\cmidrule(r){7-9}
&  &  &  & {} &   &  P & S & Q &  & {} & &\\
\midrule
\cite{Li2020survey} & X & \checkmark & X  & \checkmark &  \checkmark &  X & X & X & X & X & \checkmark & X\\
\cite{Lim2020survey}& X  & \checkmark & X & \checkmark & X & X  & X  & X &  X &\checkmark & \checkmark & X\\
\cite{Kairouz2019} & X & X & X  & X &  X &  X & X & X & \checkmark & X & \checkmark & X\\
\cite{Wahab2021survey} & X & X & X  & X & \checkmark &  X & X & X & \checkmark & X & \checkmark & X\\
\cite{Rahman2021survey}& X  & \checkmark & X & \checkmark & X & X  & X  & X &  X &\checkmark & \checkmark & X\\
\cite{Bellavista2021survey}& X  & \checkmark & X & \checkmark & X & X  & X  & X &  X &\checkmark & \checkmark & X\\
\cite{Gao2022survey} &  \checkmark  & X & X & \checkmark & X & X  & X  & X &  X & X & X & X\\
\cite{Ye2023survey}& \checkmark  & \checkmark & \checkmark & \checkmark & \checkmark & X  & X  & X &  X &\checkmark & \checkmark & X\\
\cite{Gecer2024survey}& X  & \checkmark & X & \checkmark & X & X  & X  & X &  X &\checkmark & \checkmark & X\\
Ours & \checkmark  & \checkmark & \checkmark & \checkmark & \checkmark & \checkmark  & \checkmark  & \checkmark &  \checkmark &\checkmark & \checkmark & \checkmark\\

\bottomrule
\end{tabular}
\end{adjustbox}
A1: Data Moderation, A2: Personalization, A3:Cluster, B1: Client Selection, B2: Model Compression, B3: Communication Rounds, P: Pruning, S: Sparcification, Q: Quantization

\end{table}

Numerous surveys of federated learning in general or particular facets have been made available. Still, there is a need to offer a sensible and thorough taxonomy of the current state of energy-efficient (green) federated learning research difficulties and state-of-the-art, which is a significant and developing area of research for green-aware AI. Thus, we highlight the distinctive contributions of our work in Table \ref{tab:relatedsurvey} and evaluate the important contributions and limitations of previous relevant studies. The difficulties posed by federated learning are examined by Li \etal \cite{Li2020survey} from four angles: statistical heterogeneity, system heterogeneity, communication efficiency, and privacy concerns. They also provide a quick overview of a number of potential future study areas. In the study of federated learning in mobile edge networks, Lim \etal \cite{Lim2020survey} separated the approaches now in use into two categories: those that address the core issues with federated learning and those that apply federated learning to address edge computing issues. In addition to discussing current developments in federated learning, Kairouz \etal \cite{Kairouz2019} offered an overview of unresolved issues and concerns, such as effective communication, privacy protection, attack defense, and federated fairness. Wahab \etal \cite{Wahab2021survey} offer a detailed categorization system of current problems and solutions. The work mainly considered client selection and communication efficiency approaches. Rahaman \etal \cite{Rahman2021survey} divided the survey's categories into data partitioning, FL architectures, algorithms/aggregation techniques, and personalization techniques as the implementation details and FL applications in various industries and domains, and communication cost, statistical heterogeneity, systems heterogeneity, and privacy/security as the main challenges.
 In \cite{Bellavista2021survey}, the authors included information on the characteristics of federated environments, a unique taxonomy of decentralized learning methodologies, and a thorough explanation of the most pertinent and particular system-level contributions made by the surveyed solutions for non-IIDness, privacy, device heterogeneity, and poisoning defense. Data space, statistics, systems, and model heterogeneity in FL are covered by Gao \etal \cite{Gao2022survey}, who also offer a taxonomy and an overview of the scenarios, objectives, and techniques associated with each heterogeneity challenge. In addition to offering multiple viewpoints to show the potential for future development of heterogeneous federated learning, the authors of in \cite{Ye2023survey}  developed a new taxonomy to classify current heterogeneous federated learning approaches into three levels: data-level, model-level, and server-level. Gecer \etal \cite{Gecer2024survey} conducted a survey of FL concerning efficiency, efficacy, privacy, robustness, and fairness. In order to compare and characterize federated learning solutions, the authors identified three abstract roles to present their conceptual model: aggregator, learner, and data creator. Some of the previous survey works \cite{Ma2023clientselection, Pfeiffer2023survey, Qi2024survey} are related to specific concerns such as non-IID data, heterogeneity, model aggregation etc.
\par
As evidenced in Table \ref{tab:relatedsurvey}, no state-of-the-art survey to green FL addresses energy efficiency concerns while maintaining the trade-off between communication rounds, convergence, and accuracy. Whatever the architecture or type of FL models for IoT applications, it is necessary to analyze the impact of energy efficiency and carbon footprint for a sustainable environment and green-aware AI. Moreover, this systematic survey answers the following research questions:
\begin{itemize}
\item[\textbf{RQ1}:] How FL become green-aware?
\item[\textbf{RQ2}:] What are the state-of-the-art green FL methods in the IoT application?
\item[\textbf{RQ3}:] What are the challenges affecting the green FL in IoT 
applications?
\item[\textbf{RQ4}:] What are the research gaps and potential future research directions of green FL related to IoT applications?

\end{itemize}
The primary research question (RQ1) focuses on popularly used FL variants and the roadmap of green FL. The RQ1 is addressed by sections \ref{sec:challenges} and \ref{sec:variants}.
The second research question (RQ2) is to offer a thorough and organized summary of all literature pertaining to green FL. Additionally, (RQ2) intends to demonstrate how adopting FL might be advantageous for the IoT application. The third research question (RQ3) is also driven by the need to address issues with energy efficiency and optimization in green FL IoT contexts. The RQ2 and RQ3 are addressed by \ref{sec:greenFL}. The fourth research question (RQ4) offers final recommendations for a researcher in the green FL area that are mostly focused on issues with IoT applications. RQ4 is addressed by the sections \ref{sec:application} and \ref{sec:futurescope}.
In this survey work, we explore  FL approaches and the problems of an energy-efficient optimized distributed model, as well as challenges and opportunities with the application of the Internet of Things (IoT). It is pertinent to mention that due to ample evidence of security and privacy-preserving challenges and solutions, in this survey, we are not considering the security and privacy-preserving issues.

\subsection*{Main Contributions}
The main contributions are as follows:
\begin{enumerate}
    \item we discuss in detail the challenges of Federated Learning with the solutions towards green FL.
    \item we shed light on the energy-efficient algorithms and methodologies employed in federated learning systems, reducing carbon footprints.
    \item we examine the impact of FL optimization techniques for resource allocation, minimizing energy consumption during model training across distributed devices. 
    \item Finally, we outline the difficulties, recommendations, and insights we've learned by examining the trade-offs between performance, energy efficiency, and training time in a real-world FL system.
    \end{enumerate}

We are sincerely convinced that our paper's results will provide valuable insights into the intersection of sustainability and federated learning, paving the way for eco-friendly innovations in AI and contributing significantly to the advancement of sustainable AI technologies.
\par
\begin{figure}
    \centering
    \includegraphics[scale=0.18]{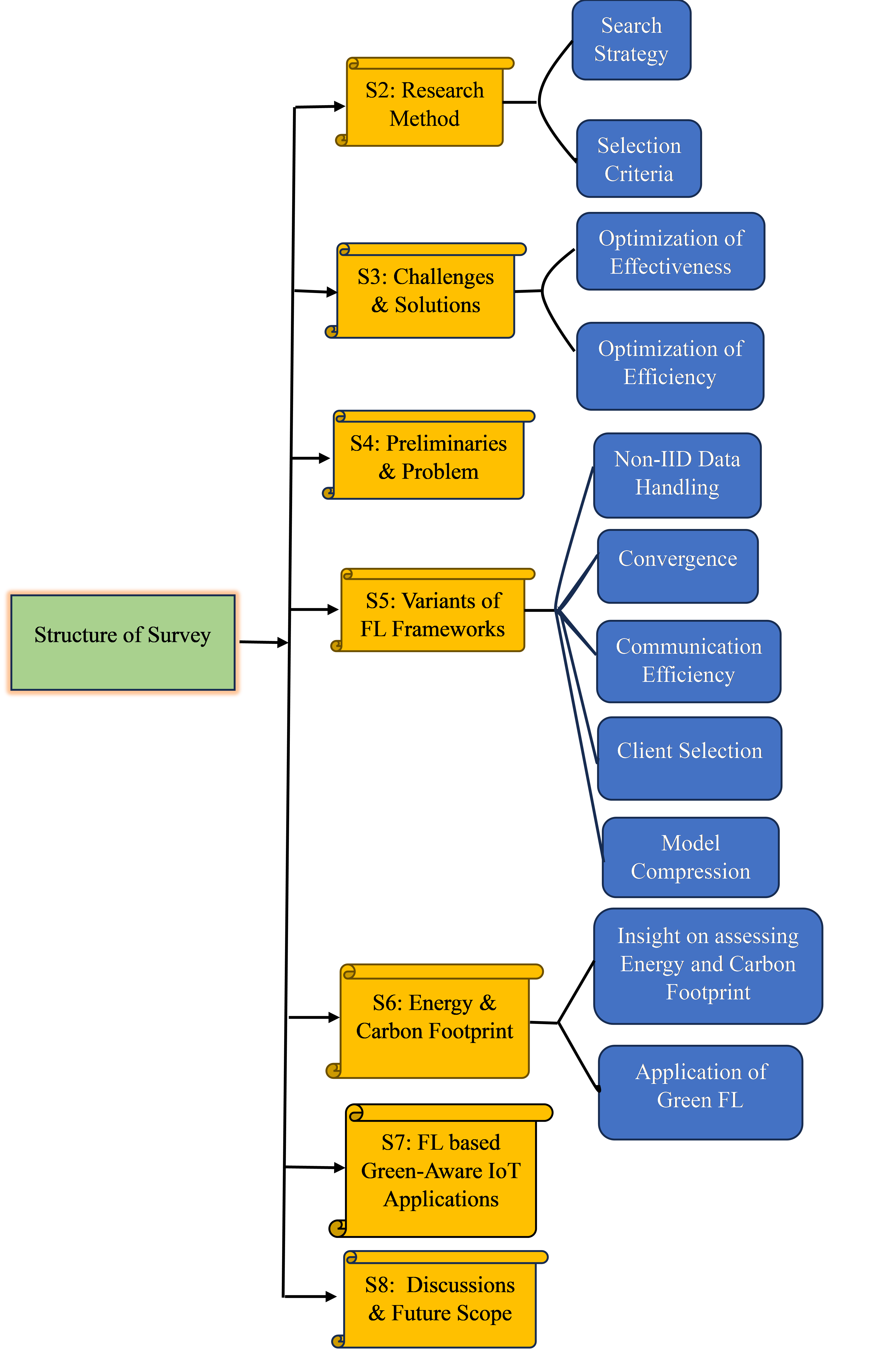}
    \caption{Structure of the Survey}
    \label{fig:structure}
\end{figure}

Figure \ref{fig:structure} shows the structure of the survey. 

\section*{Research Method}
A comprehensive literature review was carried out to evaluate the applications and impacts of green FL in Green-aware AI. This entails systematically locating and evaluating pertinent papers in the FL area of interest, ensuring that the review procedure is transparent, scalable, and scientific. Regulatory concepts are utilized to ensure a comprehensive and effective systematic review of the literature.

\begin{table}[ht]
    \begin{center}

    \begin{tabular}{p{3cm} p{7cm} p{3cm}}
   
    \hline
        Scientific Database &  Query & Studies Results\\
        \hline
         PubMed & ((green federated learning OR federated learning AND ((fft[Filter]) AND (english[Filter]) AND (2018:2024[pdat]))) AND (IoT OR IoMT AND ((fft[Filter]) AND (english[Filter]) AND (2018:2024[pdat])))) AND (``energy efficiency'' OR ``carbon emission'' OR ``carbon footprint'' OR optimization AND ((fft[Filter]) AND (english[Filter]) AND (2018:2024[pdat]))) AND ((fft[Filter]) AND (english[Filter])) AND ((fft[Filter]) AND (english[Filter])) & 21\\
         \hline
         Web of Science &``green federated learning'' OR ``federated learning'' OR ``green AI'' AND
``IoT OR IoMT'' AND ``energy efficiency OR carbon emission OR carbon footprint OR optimization'' & 20\\
\hline
IEEE Xplore & (``All Metadata'':Green federated learning) OR (``All Metadata'':federated learning) OR (``All Metadata'':green AI) AND (``All Metadata'':IoT) OR (``All Metadata'':IoMT) AND (``All Metadata'':energy efficiency) OR (``All Metadata'':carbon emission ) OR (``All Metadata'':carbon footprint) OR (``All Metadata'': optimization) & 112\\
\hline
Science Direct & (``green federated learning'' OR ``federated learning'' OR ``green AI'') AND  (IoT OR IoMT) AND (``energy efficiency'' OR ``carbon emission'' OR ``carbon footprint'' OR optimization) &325\\
\hline
ACM Digital Library & [[All: "green federated learning"] OR [All: "federated learning"] OR [All: "green ai"]] AND [All: "IoT" "IoMT"] AND [[All: "energy efficiency"] OR [All: "carbon emission"] OR [All: "carbon footprint"] OR [All: optimization]] AND [E-Publication Date: (01/01/2018 TO 31/12/2024)]&449\\
\hline
    \end{tabular}
    \end{center}
    \caption{Full Query term in publication databases}
    \label{tab:searchstrategy}
\end{table}

The research methodology used for this study is the ``preferred reporting item for systematic reviews and meta-analyses'' (PRISMA) \cite{Page2021}. The PRISMA technique is a widely used standard for providing evidence in systematic reviews, and it has been embraced by journals and organizations relevant to health. PRISMA methodologies provide several benefits, including enhancing the quality of the review, allowing the reader to evaluate its merits and shortcomings, simulating review procedures, and structuring and formatting the review using PRISMA headings. However, conducting a comprehensive examination and releasing it can take some time. It can also become rapidly obsolete, so it needs to be updated regularly to include all new primary data released since the start of the project. 
\begin{figure}
    \centering
    \includegraphics[scale=0.18]{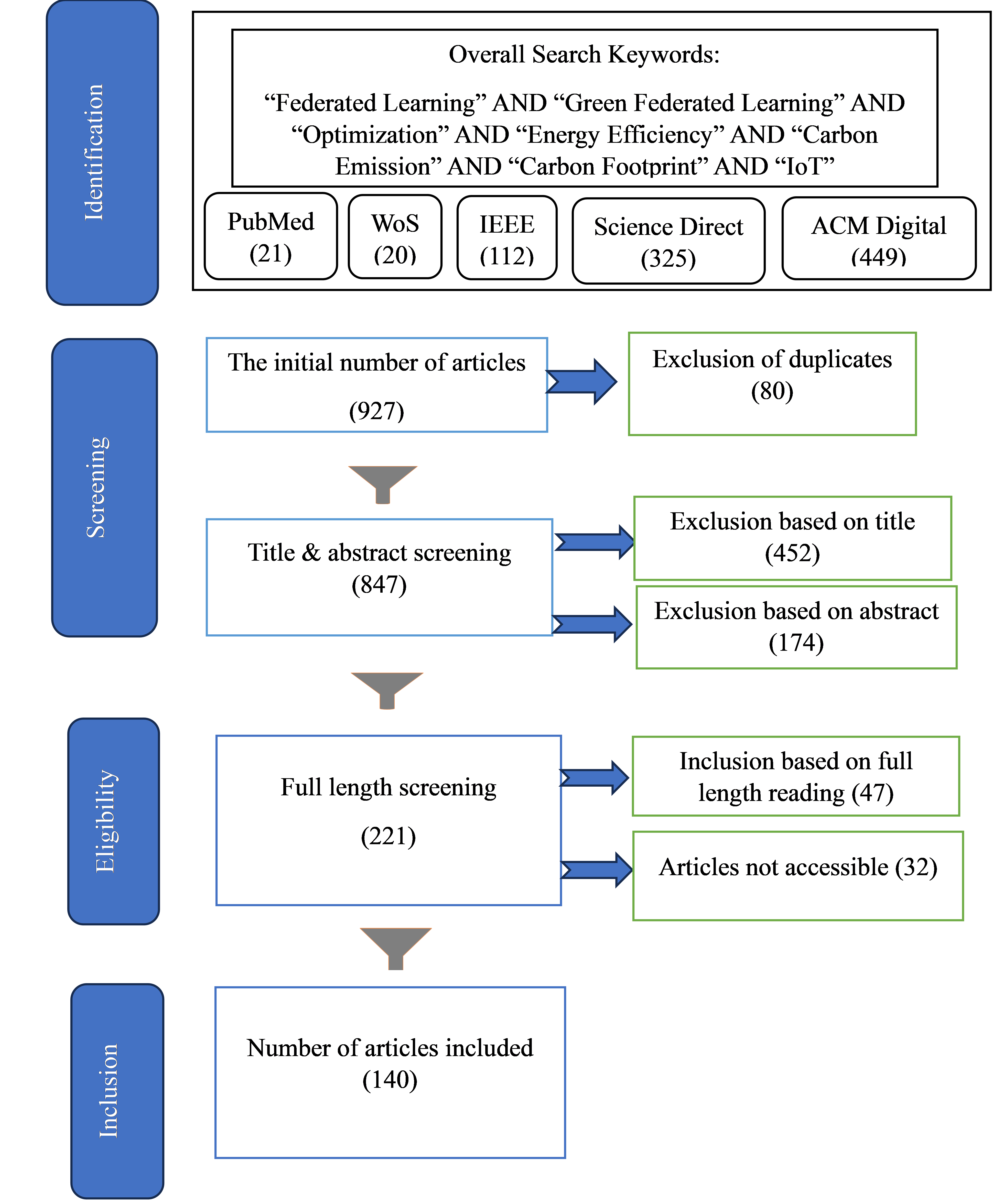}
    \caption{Challanges and potential solutions in federated Learning towards Green Federated Learning}
    \label{fig:selectioncriteria}
\end{figure}
\subsection*{Search Strategy}
A selection of terms from the specified research questions is used to create the search string. The following steps are taken to arrange the keywords using the boolean syntax "AND" in a specific order to create the search query that results in the Table \ref{tab:searchstrategy}. Based on this search strategy, we select the papers between the years 2018-2024, that are related to various challenges of FL and their respective solutions towards green FL.

\subsection*{Selection Criteria}
For scholarly research, this review only used credible and trustworthy foreign sources. Figure \ref{fig:selectioncriteria} shows the four steps of the PRISMA methodology—identification, screening, eligibility, and inclusion—in the process of choosing literature. The sources are acquired from several databases, shown in Table \ref{tab:searchstrategy}.

\section*{Challenges and solutions in Federated Learning}
\label{sec:challenges}
\begin{figure}
    \centering
    \includegraphics[scale=0.18]{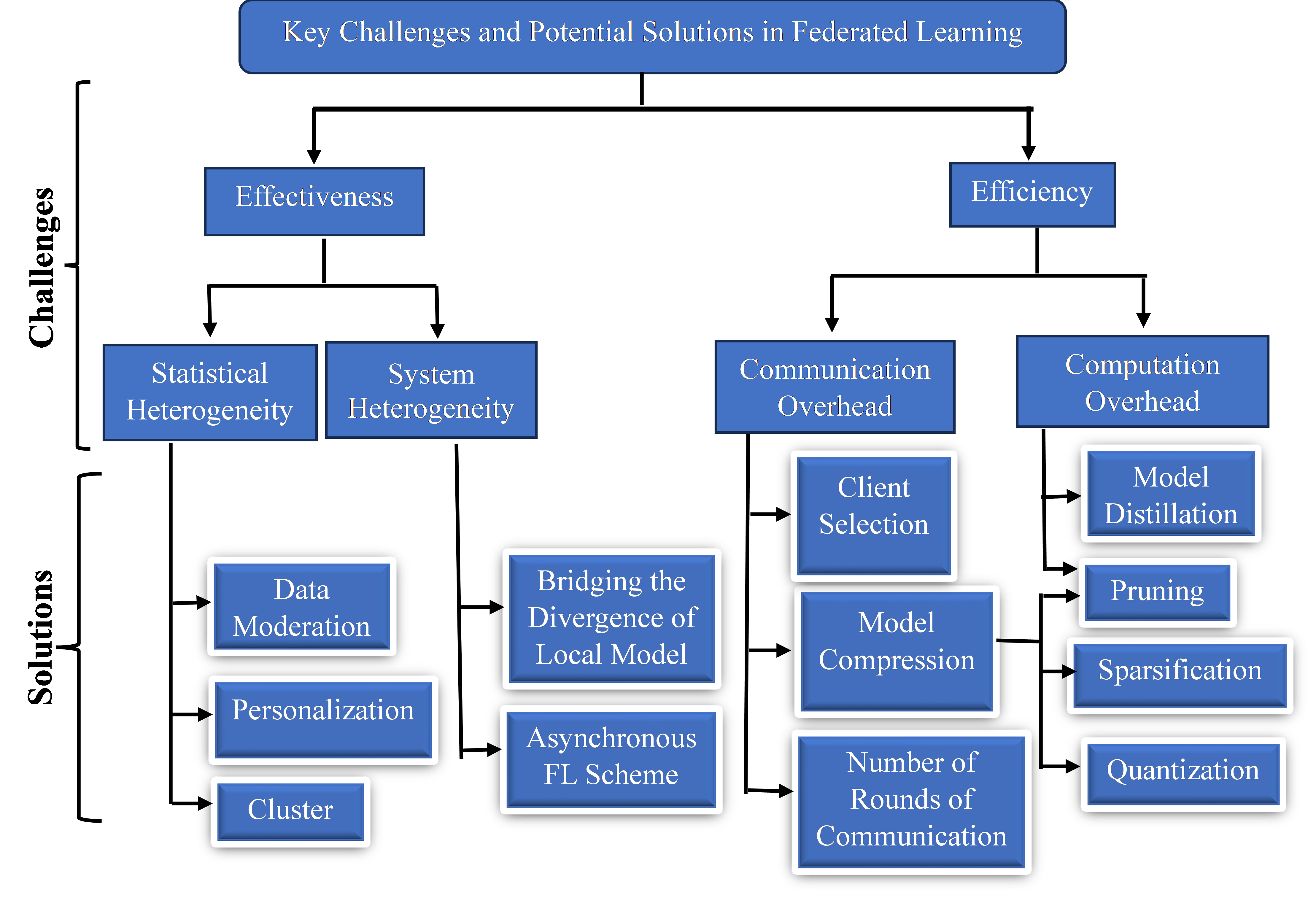}
    \caption{Challenges and potential solutions in federated Learning towards Green Federated Learning}
    \label{fig:challenges_FD}
\end{figure}
Several aspects of challenges are involved in the design of federated learning. However, in this paper, we discuss challenges directly related to energy-efficient federated learning, i.e. for Green Federated Learning (GFL). 
\par
The taxonomy of FL issues is summarized in Figure \ref{fig:challenges_FD}, where we also provide an overview of the essential solutions (which will be covered in more detail in the upcoming sections). However, privacy and security challenges are not considered in this work. In the FL literature, an adequate number of papers are available regarding the privacy and security challenges with the respective solutions. 
\par

\subsection*{Optimization of Effectiveness} \label{subsec:effectiveness}

\textit{Effectiveness} is obtaining satisfactory global (and local) models that consider different client heterogeneity.
The efficacy of FL is significantly impacted by two factors: system heterogeneity and statistical heterogeneity. 

1) System heterogeneity: System parameters like hardware, network connectivity, and energy limitations frequently differ between different clients \cite{Wei2022vertical}. The straggler effect may result from this, namely in cross-device FL.

2) Statistical heterogeneity is related to client non-i.i.d. data \cite{Zhu2021noniid}. For instance, different hospitals could have non-identifiable disease data because their patient population varies in geography and demographics. Different banks may have heterogeneous data because clients from various areas and nations may have varied financial backgrounds and investing styles.
Three key approaches have been created by current studies to address statistical heterogeneity, and they are presented as follows:\\
\textbf{Data Moderation}: Predictably, the heterogeneous data distributions cause the FL performance loss with non-i.i.d. data. Data moderation aims to address this problem by altering the distributions. Data sharing is one potential technique wherein clients' local models are trained using local and shared global data at the server. Data augmentation is an additional technique that adds augmentations to the local data based on information from data distributions of other clients. \\
\textbf{Personalization}: Due to the drift of heterogeneous data, the global model generated by the server may hurt the performance and generalization of the local models on the clients. To solve this problem, personalization techniques modify the local models according to the local tasks. Various techniques are available for personalization. The approaches to this kind of personalization are either fine-tuning a global model on a local client distribution or pruning a global model. Nevertheless, these current approaches either predetermine network layers for fine-tuning, resulting in inadequate global information storage inside client models, or personalize at the expense of critical global knowledge retention \cite{Tamirisa2023fedselect}. Another approach for personalization is personalization layers. The negative consequences of statistical heterogeneity can be mitigated using personalization layers not subject to the federated averaging (FedAvg) process \cite{Arivazhagan2019federated}. Another approach for personalization is knowledge distillation, a technique for knowledge transfer, to compare the results of a shared dataset and learn from each other's inferred knowledge \cite{Jeong2023personalized}.\\
\textbf{Clustering}: Though most FL frameworks operate under the assumption that there is only one global model, learning clients' knowledge may not be adequate in extremely heterogeneous data sets. Clustering techniques address this problem by dividing the clients into several clusters and ensuring that each cluster upholds a server-client architecture. Specifically, similar clients are assigned to the same cluster based on the loss values or submitted model weights that indicate how similar they are \cite{Ouyang2021clusterFL}.

\subsection*{Optimization of Efficiency}
Federated learning strives to reduce computing and communication overheads while addressing the efficiency challenge. 
\subsubsection*{Communication Overhead}
The literature mentions two main approaches to reduce communication overhead in federated learning. They are model compression and communication rounds \cite{Morell2022optimising}. Client selection is another approach to reduce communication overhead as the involvement of more client participation makes convergent the global model earlier. Reducing the number of global iterations by synchronizing the maximum number of clients eventually lowers communication expenses and energy usage. \cite{Thakur2023FL}.\\
\textbf{Model compression} can be achieved either using quantization or sparsification. Sparsification significantly lowers communication overhead, but it must be used carefully since transmitting too little information might negatively impact accuracy, while transmitting too much information can result in energy depletion and network congestion. Consequently, there is a trade-off between excessive and insufficient compression. It is costly to try to find the ideal balance through hyperparameter tweaking, and the tuning parameters that are produced are static and problem-specific. Furthermore, the majority of current compression techniques don't take into account the varying communication costs associated with various technologies \cite{Khirirat2021}. Furthermore, because edge devices in the Internet of Things (IoT) have high energy constraints, the compression framework should be created to guarantee that the greatest amount of learning can occur while utilizing the least amount of energy to transmit the models—that is, to maximize the energy efficiency of edge nodes.\\
\textbf{Sparsification} is a model compression technique for automatically determining the near-optimal communication and computation trade-off \cite{Han2020adaptive}. This trade-off is adjusted in the original FL method, federated averaging (FedAvg), by the number of local update rounds that occur in between each of the two communication (weight aggregation) rounds \cite{McMahan2016}. FedAvg either delivers all of the model parameters or nothing at all following each local model update step using gradient descent. Gradient sparsification (GS), a more balanced method that sends a sparse vector containing a selection of significant values from the complete gradient, has drawn interest recently in remote learning systems \cite{Jiang2018}.
In contrast to FedAvg's "send-all-or-nothing" strategy, GS offers more flexibility in managing the trade-off between computation and communication \cite{Han2020adaptive}. However, these approaches employ a single compression rate for every user and fail to take into account the communication heterogeneity in an actual FL system; consequently, these approaches are constrained by the worst communication capacity among users. Moreover, sparsification approaches lack adaptability and fail to leverage the redundant, similar information in users' ML models for compression. In \cite{Beitollahi2022}, the authors presented a novel approach to Dynamic Sparsification for Federated Learning (DSFL) that allows users to compress their local models according to their communication capacity at each iteration through the use of two novel methods for sparsification: layer-wise similarity sparsification (LSS) and extended top-K sparsification. DSFL is made possible by LSS to take advantage of the global redundant information present in users' models through the sparsification process of Centralized Kernel Alignment (CKA) similarity. By allowing various values of the sparsification rate K for each user at each iteration, the extended top-K model sparsification approach enables DSFL to accommodate the diverse communication capability of user devices. Still, maintaining accuracy following a high ratio sparsification has never been addressed. To rectify the sparse gradient update process, a General Gradient Sparsification (GGS) framework is put forth in \cite{Li2020Spercification} for adaptive optimizers. Gradient correction and batch normalization update with local gradients (BN-LG) are its two main techniques. The optimizer can correctly handle the accumulated unimportant gradients with gradient correction, improving the model's convergence. Moreover, local gradient updates to the batch normalization layer can mitigate the effects of delayed gradients without adding to the communication overhead. \\
\textbf{Quantization} is another widely used model compression technique. The FL setup presents several difficulties. For instance, depending on the requirements for their hardware design, various clients may have varying computational limitations. The hardware accelerator's supported quantization bit-width is one realistically significant heterogeneous hardware attribute. A well-trained model ought to be quantized to a range of bit widths that are representative of the heterogeneous device landscape without exhibiting appreciable performance reduction. The goal of quantization robustness is to do this by training a single model with the constraint of robustness to different quantization bit-widths while at inference time, without the need for retraining or fine-tuning \cite{Gupta2023quantization}.
The research hasn't given enough thought to the energy requirements and hardware-induced limits for on-device learning, despite the effectiveness of common Federated optimization techniques like Federated Averaging (FedAvg) in FL. Enabling trained models to be quantized to different bit-widths based on the energy requirements and heterogeneous hardware designs across the federation is specifically a requirement for on-device learning. 
\par
Nevertheless, these approaches, such as sparsification and quantization are pre-defined. They fail to capture redundant, correlated information across FL iterations, user device data, and ML model parameters. Furthermore, the error-correcting potential of the FL process is not entirely utilized by these methods. In \cite{Beitollahi2022Globecom}, the authors presented the Federated Learning with Autoencoder Compression (FLAC) method, which compresses user devices' models for uplink transmission by taking advantage of FL's redundant information and error-correcting capabilities. In the Training State, FLAC trains an autoencoder on the server to encode and decode user models. Later, in the Compression State, the autoencoder is sent to user devices to compress local models for further iterations. 
FLAC dynamically manages the autoencoder error by alternating between the Training State and Compression State to modify its autoencoder and its compression rate based on the error tolerance of the FL system, ensuring the convergence of the FL.
\\
\textbf{Client Selection} is another method to address the communication efficiency challenge in FL. Transferring a large number of updated model parameters over the uplink communication channel from the users to the server is a challenging task in FL which increases the communication complexity is a challenging task. Reducing the number of participating clients by scheduling policies \cite{Amiri2020, Yang2020} can help address this difficulty. In the literature, only a subset of clients can be scheduled for updates at each iteration due to bandwidth limitations. However, the random selection of a partial number of clients can reduce the convergence speed in the presence of heterogeneous clients. Moreover, selecting more clients ensures fast convergence in the presence of heterogeneous clients \cite{Huang2023clientselection}. Hence, it is necessary to select participating clients to maintain the tradeoff between accuracy, convergence speed, and communication efficiency.

\subsubsection*{Computation Overhead}
Computational resources, such as CPU cycles and energy consumption, are used during the local model training procedure. Choosing appropriate and potentially simpler local model structures (e.g., switching from deep neural networks to random forests) is a feasible way to lower the processing overheads. This can be challenging because there isn't a single accepted standard for selecting models, and selecting the right or ideal model structure is still up for debate in the field of machine learning.

\section*{Preliminaries and Problem Definition}
FL has a server-client architecture akin to distributed stochastic gradient descent (SGD), in which client workers or devices perform training and exchange knowledge with a central server. The goal of FL's simplified scenario is to develop a model while keeping in mind that training data is dispersed locally across numerous devices. Therefore, by determining the ideal NN weights $\omega$ s.t., the objective is to minimize the following function.
\begin{equation}
\begin{aligned}
\min_{\omega} L(\omega) \quad & \textrm{where} \quad & L(\omega)=\frac{1}{|C|}\sum_{c \in C} L_c(\omega)\\
\end{aligned}
\label{equ:minloss1}
\end{equation}
where $L_c(\omega)$ is the loss function of device $c$, where $c$ is a device inside the set $C$, and $L(\omega)$ is the global loss (at a centralized server that handles parameter aggregation). Only the local dataset $D_c = {x_c,y_c}$, where $x_c$ is the input, and $y_c$ is the label, is accessible to each device only. Therefore, the function $l_c(\omega)$ can be rewritten as
\begin{equation}
\begin{aligned}
 L_c(\omega)=l(x_c,y_c,\omega) \quad \{x_c,y_c\} \sim \mathcal{P} \enspace \forall c\\
\end{aligned}
\label{equ:minloss2}
\end{equation}
The distribution $\mathcal{P}$ provides samples for each device $c$, which leads to $|C|$ disjoint partitions of the entire dataset. Two bounds on the correctness of such a scheme are as follows. First, the centralized training scenario, in which a device has access to the entire dataset, is a natural upper accuracy bound. By depending solely on its local data, the second device lacks knowledge transfer and naturally lowers the accuracy. Two primary objectives of computational green-aware FL methods are ascertained: despite having limited devices, increasing the convergence speed with less energy and carbon footprint and achieving a high final accuracy.

\section*{Variants of Federated Learning Frameworks}
\label{sec:variants}

Several FL approaches are proposed in the literature for various applications to address the challenges of FL, which differentiate the FL from traditional distributed optimization. The state-of-the-art methods are divided into four aspects based on effectiveness and efficiency. First is the non-IID data handling. The second is related to selecting an optimal number of clients to improve the performance of FL in terms of convergence rate and model accuracy. The third is to compress deep neural network models to reduce energy expenditure. The fourth is to minimize communication overhead. According to the literature, the green FL should be communication- or energy-efficient.
\subsection*{Non-IID Data Handling FL}
A new distributed machine learning platform for protecting privacy is called federated learning. However, when the training data are not independently and identically distributed (Non-IID) on the local devices, models built in FL typically perform worse than those trained in the conventional centralized learning mode \cite{Zhu2021noniid}. To handle this issue, several authors have proposed different FL approaches. In \cite{McMahan2016}, the authors proposed a simple FL algorithm, \textbf{FedAvg}, to handle non-IID data. The authors' comprehensive testing of this method demonstrates that it can significantly reduce the number of communication rounds needed to train a deep network using decentralized data. It is resilient to non-IID and imbalanced data distributions. FedAvg trains high-quality models with very few communication rounds, as demonstrated by results on a variety of model architectures, such as a multi-layer perceptron, two different convolutional NNs, a two-layer character LSTM, and a large-scale word-level LSTM.
 However, the authors in \cite{Sahu2018FederatedOI} claimed that the FedAvg failed to give the convergence guarantee with non-IID data and addressed heterogeneity in federated networks by introducing the \textbf{FedProx} framework. However, FedProx can be seen as a generalization and re-parametrization of FedAvg. Although the approach is only slightly altered by this re-parameterization, the changes have significant implications for theory and practice. FedProx is guaranteed to converge theoretically when learning over data from non-identical distributions (statistical heterogeneity) and when complying with device-level systems constraints by letting each participating device do a different amount of work (systems heterogeneity). In practice, on various real-world federated datasets, FedProx enables more robust convergence than FedAvg. Specifically, FedProx shows far more accurate and persistent convergence behavior than FedAvg, enhancing absolute test accuracy by 22\% on average in extremely heterogeneous situations. At the same time, in \cite{Zhao2018noniid}, the authors experimentally demonstrated that when neural networks are trained with highly skewed non-IID data, where each client device trains just on a single class of data, the accuracy of federated learning (FedAvg) decreases significantly by as much as ~55\%. To address this, the authors suggested \textbf{FedCurv} \cite{Shoham2019fedcurv}, an FL-based method that enhances training on non-IID data: developing a limited subset of data that is globally shared among all edge devices. In \cite{Karimireddy2020noniid}, the authors demonstrated that when the data is heterogeneous (non-iid), FedAvg experiences ``client-drift,'' which leads to unstable and sluggish convergence. In order to address this, the authors in \cite{Karimireddy2020noniid} suggested a brand-new technique called \textbf{SCAFFOLD}, which makes use of control variates (variance reduction) to adjust for ``client-drift'' in local updates. SCAFFOLD is insensitive to client sampling and data heterogeneity, requiring a notably smaller number of communication rounds. Additionally, SCAFFOLD can benefit from data similarity in the client, resulting in even faster convergence for quadratic. \textbf{FedPer} \cite{Arivazhagan2019FedPer} is a personalized FL framework which also handles the non-i.i.d. data in an efficient manner. 

\begin{table}
    \setlength\tabcolsep{2pt} 
\footnotesize  
\centering
    \caption{Variants of FL Algorithms}
    \label{tab:FL_models} 
    \resizebox{\textwidth}{!}{
   \begin{tabular}{|c|c|c|c|c|c|}
   \hline
    \textbf{FL Model} & \textbf{Heterogeneity} & \textbf{Client Selection} & \textbf{Communication Round} & \textbf{Convergence} & \textbf{Energy Efficiency}\\
    \hline
    FedAvg \cite{McMahan2016} & X & X & X & X & X  \\
    \hline
    FedProx \cite{Sahu2018FederatedOI} & \checkmark & X & X & X & X \\
    \hline
  FedPer \cite{Arivazhagan2019FedPer}  & \checkmark & X & X & X & X \\
    \hline
    
     FedCurv \cite{Shoham2019fedcurv} & \checkmark & X & X & \checkmark & X \\
    \hline
    SCAFFOLD\cite{Karimireddy2020noniid} & \checkmark & X & \checkmark & \checkmark & X \\
    \hline
    MOON \cite{Li2021MOON} & \checkmark & X & X & \checkmark & X \\
    \hline
    FedNova \cite{Wang2020FEDNOVA}& \checkmark & X & X & \checkmark & X \\
    \hline
    FedPAQ \cite{Reisizadeh2019FedPAQAC} & X & X & \checkmark & \checkmark & X\\
    \hline
    FEDEMD \cite{Huang2023clientselection} & \checkmark & \checkmark & X & \checkmark & X \\
    \hline
    FedDyn \cite{Acar2021FedDyn}& \checkmark & X & \checkmark & X & X \\
    \hline
    FedMA \cite{Wang2020FedMA} & \checkmark & X & \checkmark & X & X \\
    \hline
    FedPD \cite{Zhang2020FedPDAF} & \checkmark & X & \checkmark & \checkmark & X \\
    
    \hline
    \end{tabular}}
\end{table}

\subsection*{Convergence}
\textbf{FedNova} \cite{Wang2020FEDNOVA} is a quick error convergence normalized averaging method that removes objective inconsistency. This method mainly focuses on the heterogeneity of local updates of the model. It gives the first systematic understanding of the solution bias and the convergence delay caused by objective inconsistency, and it incorporates previously presented methods like FedAvg and FedProx. \textbf{SCAFFOLD} \cite{Karimireddy2020noniid} is another proposed FL algorithm that handles both the non-IID data and convergence rate. The authors in \cite{Karimireddy2020noniid} proved that the FedAvg FL algorithm suffers from client drift while using heterogeneous (non-i.i.d) data with stochastic Gradient Descent (SGD) as a local solver, resulting in slow convergence. The authors offered SCAFFOLD as a remedy that accounts for the client drift in its local updates using control variates (variance reduction). Moreover, SCAFFOLD is insensitive to client sampling and data heterogeneity, requiring a notably smaller number of communication rounds. \textbf{Model-contrastive learning (MOON)} \cite{Li2021MOON} is another FL method that can handle non-i.i.d data and convergence rate simultaneously. MOON is a unique approach to addressing the non-IID data issue and a straightforward and efficient FL framework.

\subsection*{Communication-efficiency}
The authors in \cite{Reisizadeh2019FedPAQAC} proposed a communication-efficient FL method known as \textbf{FedPAQ}. FedPAQ is based on periodic averaging and quantization to overcome the communication bottleneck and scalability challenges. 
FedPAQ is based on three main features: (1) partial device participation, where only a portion of the devices take part in each training round; (2) periodic averaging, where models are updated locally at devices and only periodically averaged at the server; and (3) quantized message passing, where the edge nodes quantize their updates before uploading to the parameter server. These aspects address the scalability and communication issues in federated learning. We additionally illustrate the near-optimal theoretical guarantees that FedPAQ achieves for both strongly convex and non-convex loss functions and the empirical tradeoff between computation and communication. \textbf{Federated Matched Averaging (FedMA)} \cite{Wang2020FedMA} is another FL approach that reduces communication rounds efficiently with heterogeneous data. For contemporary CNNs and LSTMs, FedMA is a novel layers-wise federated learning approach that leverages Bayesian nonparametric techniques to adjust to data heterogeneity. The authors provided empirical evidence demonstrating that FedMA outperforms the most advanced FL algorithms while also reducing the overhead of communications. \textbf{FedDyn} \cite{Acar2021FedDyn}, a dynamic regularization FL technique. A novel idea at the core of FedDyn is the dynamic updating of each device's risk objective in each round to guarantee that the device optima are asymptotically compatible with the stationary points of the global empirical loss. The authors obtained sharp findings for communication rounds necessary to achieve target accuracy, and they proved to have convergence results for FedDyn in both convex and non-convex scenarios. The convex case results significantly outperform previous state-of-the-art efforts. In benchmark examples, FedDyn consistently achieves considerable communication savings over competing techniques on massively distributed large-scale text data, spanning different options of device participation and heterogeneity. The quality of communication lines, heterogeneity, and massively distributed data do not theoretically affect FedDyn.
\textbf{Federated Primal-Dual (FedPD)} \cite{Zhang2020FedPDAF} attains convergence under specific assumptions and the best possible optimization and communication complexity when the data is not i.i.d.  

\par
All the above-mentioned FL approaches in Table \ref{tab:FL_models} take care of either non-i.i.d. data, prove the convergence with non-iid data, or prove the convergence with non-iid data and reduce the communication round—however, none of the discussed about the energy efficiency, which is a crucial problem with IoT devices. Therefore, the researchers started thinking about energy-efficient FL approaches. Some researchers claimed that the energy efficiency of the FL framework is directly related to the number of participating clients, the number of communication rounds, and the low-precision model. In the previous section, we discussed several FL algorithms that claim communication efficiency, i.e., the algorithms converge with a minimum number of rounds. The further sections demonstrate the impact of the number of participating clients and the low-precision model.
\begin{table}
    
\centering
    \caption{Client selection FL Algorithms}
    \label{tab:FL_clientselection} 
    \resizebox{\textwidth}{!}{
   \begin{tabular}{|l|l|l|p{9cm}|}
   
  \hline
    \multirow{2}{*}{FL Model/Reference} & \multicolumn{2}{l|}{Heterogeneity} & \multirow{2}{*}{Description}\\\cline{2-3}
    
   {}   & Data   & System    & {}\\
   \hline
   FedCS \cite{Nishio2018ClientSF} & X &\checkmark & Before a deadline, select as many clients as you can.  It is predicated on a knapsack-constrained greedy method.\\
   \cite{Elsa2021} &\checkmark & X & Two-tier significance sampling for both data and clients. Clients are chosen initially, and then their data is sampled.\\
   OCEAN \cite{Xu2021} & X & \checkmark & Allocation of bandwidth based on energy limits of the client. In order to improve the client selection pattern, it makes use of wireless channel information.\\
   Oort\cite{Oort-osdi21} & \checkmark & \checkmark & Make use of client-side heterogeneous systems and data. In order to choose participants based on their resilience to outliers, it uses an exploration-exploitation technique. \\
   \cite{Amiri2021clientselection} & \checkmark & \checkmark & Learn about local gradient norm and shared block fading wireless channels. It creates a policy for allocating resources in order to plan low-profile client devices. \\
   FOLB \cite{Nguyen2021} &\checkmark & \checkmark & Based on how well the local and global updates correlate. It makes an estimate of the clients' potential to contribute to the updates. \\
   \cite{Chen2022optimalCS} & \checkmark & X & To address the communication bottleneck, employ norm-based client selection. It uses a basic algorithm to approximate the ideal client selection formula.\\
   FEDPNS\cite{Wu2022clentselection} & \checkmark & X  & Removing negative local updates through a comparison of the global and local gradients. Clients that accelerate model convergence are given preference.\\
   Power-of-Choice \cite{Cho2022clientselection} & \checkmark & \checkmark & Balance between solution bias and convergence speed in local loss-based client selection. It demonstrates how skewed client selection might hasten convergence.   \\
  PyramidFL \cite{Li2022clientselection}  & \checkmark & \checkmark & Take advantage of system and data heterogeneity within a chosen clientele. It then selects the clients depending on their utilities and optimizes utility profiling locally.\\
FCFL \cite{Zhou2022clientselection} & \checkmark & \checkmark & Wearable technology under unfavorable networking circumstances. It suggests aggregating only the model updates with the highest contributions using movement-aware FL.\\

Eiffel \cite{Sultana2022ClientSelection} & \checkmark & \checkmark & Together, take into account elements like the age of the update and resource requirements. The frequency of global and local model updates is adaptively adjusted. \\

\cite{Luo2022clentselection} & \checkmark & \checkmark & Customized client sampling probabilities are optimized to handle statistical and system heterogeneity. The wall-clock convergence time is reduced. \\

F3AST \cite{Ribero2023clientselection} & X & \checkmark &  Acquires knowledge of a client selection method that is contingent on availability in order to reduce the effect of client-selection variance on the convergence of the global model. \\

\cite{Ma2023clientselection} & X & \checkmark & 
Provide a block-wise incremental gradient aggregation approach along with a dynamic user and task scheduling scheme. \\

   \hline
    \end{tabular}}
\end{table}

\subsection*{Client-selection}
The FL process's crucial step of client selection may impact the convergence rate, model precision, and clients' overall energy usage. Moreover, the heterogeneity problem can be tackled by an efficient client selection algorithm \cite{Fu2023clientselection}. A successful FL client selection technique can strengthen robustness \cite{Nguyen2021}, increase fairness \cite{Sultana2022ClientSelection}, decrease training overheads \cite{Li2022clientselection}, and improve model accuracy \cite{Lai2021clientselection}. As a result, FL client selection research has been developing quickly in the last few years, according to the research community \cite{Soltani2022}.
Many strategies seek to choose the most significant number of clients who can complete their model training and upload before the cutoff date in each global iteration. Federated learning needs to choose the right clients at the right time in order to speed up convergence and increase accuracy. In \cite{Huang2023clientselection}, the authors provided \textbf{FEDEMD}, an adaptive client selection method based on the local and global data distributions' Wasserstein distance. When there are multiple types of Mavericks present, FEDEMD reliably assures quick convergence by adjusting the selection probability so that Mavericks are preferred when the model gains from improvement on rare classes. When compared to current methods, such as those based on Shapley values, FEDEMD enhances neural network classifier convergence by at least 26.9\% for FedAvg aggregation when compared to the state of the art. All the client selection strategies mentioned in Table \ref{tab:FL_clientselection} maintain the trade-off between communication and computation efficiency. However, reduced communication rounds and computation efficiency minimize the energy consumption of an FL model. Hence, make a green FL model.

\subsection*{Model Compression}
Model compression is proposed to minimize the neural network's complexity before it is deployed for edge devices with restricted resources. In 1988, the first studies on model pruning were conducted. An amplitude-based pruning technique for shallow, fully linked networks was presented by Hanson and Pratt \cite{Hanson1988}. Recently, in order to compress CNNs, the authors in \cite{Han2015compression} used many techniques, including quantization, pruning, and Huffman coding. Summing the convolution kernel's absolute value was suggested by Li et al. \cite{Li2017compression} as a criterion for determining its significance. This was the first instance of modifying the model structure to accomplish network compression using the convolution kernel as the pruning unit. Several criteria are proposed by researchers in \cite{Lin2020compression, Li2017compression, Liu2015compression, Liu2017learning, He2017, Molchanov2017pruning, Lee2019snip, He2019filter, Guo2020DMCP} and others to assess the role of convolution kernels in structured pruning; superfluous convolution kernels are eliminated or set to zero. To find redundant parameters during training, three pruning techniques—iterative pruning \cite{Han2015}, gentle pruning \cite{He2020}, and dynamic pruning \cite{Shen2021}—were suggested. Currently, there is no consensus on assessing the model's parameter efficacy while minimizing accuracy loss.
\newpage
\begin{center}
\begin{longtable}{|l|l|l|l|p{7cm}|}
\caption{Model Compression in Federated Learning - P:Pruning, Q: Quantization and S: Sparsification}\label{tab:FL_compression}\\
\hline
\endfirsthead
\caption* {\textbf{Table \ref{tab:FL_compression} Continued:} Model Compression in Federated Learning - P: Pruning, Q: Quantization and S: Sparsification}\\\hline
    \multirow{2}{*}{Reference} & \multicolumn{3}{l|}{Compression} & \multirow{2}{*}{Description}\\\cline{2-4}
    {}   & P   & Q & S  & {}\\\hline
   \endhead
      \multirow{2}{*}{Reference} & \multicolumn{3}{l|}{Compression} & \multirow{2}{*}{Description}\\\cline{2-4}
    {}   & P   & Q & S  & {}\\ 
     \hline
   FL-PQSU \cite{Xu2021compression} & \checkmark & \checkmark & X & It is made up of a three-stage pipeline that combines weight quantization, structured pruning, and selective updating to lower computing, storage, and communication costs and speed up FL training. \\
   FedTiny \cite{Huang2023fedtiny} & \checkmark & X & X & Produced customized small models for participating devices with sensitive local data that are limited in terms of memory and computation. Presented an adaptive batch normalization (BN) selection module to adaptively get an initially trimmed model to meet deployment circumstances, mitigating the effects of biased pruning caused by unseen heterogeneity data over devices.\\
   \cite{Shah2023compression} & X & X & \checkmark &  In order to tackle, an infeasible volume of data, compression methods to build sparse models from dense networks with much less bandwidth and storage needed. To do this, compression methods for the client models, which deal with upstream communication, and the server model, which deals with downstream communication have been considered. These two are essential for creating and preserving sparsity over communication cycles.\\ 
   FedCOMGATE \cite{Haddadpour2020compression} & X & \checkmark & \checkmark & It is a federated averaging variation for heterogeneous environments that includes compression and local gradient tracking. For generic non-convex, strongly convex, and convex objectives, it can determine its convergence rates. \\
  FedHBAA—FL \cite{Chen2024compression} & X & \checkmark & X & combined server aggregation technique with quantization bit allocation. A server aggregation technique in FedHBAA for any given client bit allotment. For both the strongly convex and nonconvex loss functions, a closed-form aggregation weight solution as a function of the allotted bits is achieved by minimizing the drift term in the convergence rate analysis. The best bit allocation plan is then found by solving the derived optimal aggregation weights.\\
  AGQFL \cite{Lian2021compression} & X & \checkmark & X & Three components make up this automatic gradient quantization method: the quantization indicator, quantization strategy, and quantization optimizer. By evaluating the current model's capacity for convergence, the quantization indicator module automatically ascertains the direction in which the quantization precision should be adjusted. The quantization strategy module modifies the quantization precision at run-time based on the indication and each node's physical bandwidth. Additionally, in order to minimize training bias and get rid of instability during the training phase, the quantization optimizer module creates a new optimizer.\\
  NQFL \cite{Chen2024NQFLcompression} & X & \checkmark & X & The significant communication overhead that comes with FL is still a major obstacle. This letter introduces a nonuniform quantization approach based on the Lloyd-Max algorithm to address this problem. This method uses fewer communication resources to accomplish the same result as other methods. Through numerical simulations and performance analysis, we confirm the convergence and efficiency of the suggested approach. It illustrates how our method can save communication overhead while preserving dependable FL system performance. \\
 FedQNN \cite{Yu2023compression} & X & \checkmark & \checkmark & A communication-and computation-efficient FL framework for Internet of Things applications. For the first time, ultralow-bitwidth quantization is integrated into the FL environment, enabling clients to carry out lightweight fix-point computing effectively and with reduced power consumption. Furthermore, by combining quantization and sparsification techniques, both upstream and downstream data are greatly reduced for more effective communication. \\
 FedDQ \cite{Qu2022compression} & X & \checkmark & X & A descending quantization system, is proposed based on the theoretical analysis that the range of the model updates will progressively contract because the model is meant to converge with training, meaning that the quantization level should decrease as training progresses.\\
 AdaQuantFL \cite{Jhunjhunwala2021AdaQuantFL} & X & \checkmark & X & An adaptive quantization technique modifies the number of quantization levels throughout training with the goal of achieving both a low error floor and effective communication. \\
 FedQOGD \cite{Park2022FedQOGD} & X & \checkmark & X & Gradient descent quantized communication efficient FL framework using partial node participation and stochastic quantization. FedQOGD has a minimal communication overhead and maintains edge-node privacy while yielding the same asymptotic performance as its centralized equivalent (i.e., all local data are gathered at the central server).\\
   \hline
\end{longtable}
\end{center}

In order to alleviate the communication delay constraint in federated learning, gradient compression—which includes gradient quantization and sparsification—has been extensively studied. Quantization modifies a floating-point number of bits to compress parameters \cite{Tonellotto2021quantization}. Low-precision representation of deep neural networks (DNNs) is a potential method to allow for effective memory reduction and acceleration. Hence, the reduction in energy consumption.

\section*{Energy and Carbon Footprint Study in Federated Learning} \label{sec:greenFL}
In the previous section \ref{sec:challenges}, several FL-related challenges and solutions are discussed elaborately. However, how FL can affect the ecosystem is still unknown and unclear. There is a relation between FL design and carbon emission \cite{Qiu2021carbon}. According to the survey, an energy-efficient FL framework reduces the carbon footprint. The potential for ML systems to greatly increase carbon emissions exists. Recent research \cite{Schwartz2019green, Strubell_Ganesh_McCallum_2020} has recommended a number of mitigating techniques and used case studies to illustrate these potential implications. To more accurately assess the overall energy and carbon footprints of machine learning—in both research and production settings—systematic and precise measurements are required. Among other advantages, accurate accounting of carbon and energy impacts promotes awareness, motivates mitigation efforts, and aligns incentives with energy efficiency \cite{Schwartz2019green}. However, energy or carbon emissions data are not routinely reported in the majority of ML research articles. Recent research primarily looks into how machine learning research affects the climate. By analyzing the anticipated power consumption and carbon emissions for a series of case studies, Strubell et al. \cite{Strubell_Ganesh_McCallum_2020} illustrated the problem of carbon and energy consequences of big NLP models. According to the authors, ``researchers should prioritize computationally efficient hardware and algorithms'', ``academic researchers need equitable access to computation resources'', and ``authors should report training time and sensitivity to hyperparameters.'' Similar recommendations are made by Schwartz et al. \cite{Schwartz2019green}, who propose floating point operations (FPOs) as a guiding efficiency parameter.
A website for calculating carbon emissions based on GPU kind, experiment duration, and cloud provider was just made available by Lacoste et al. \cite{lacoste2019quantifying}.
Similarly, FL - a cooperative machine learning method that uses data from decentralized entities to train a centralized model—can also be resource-intensive and have a significant carbon impact, mainly when used on a large scale. In contrast to centralized artificial intelligence (AI), which may reliably utilize renewable energy sources at carefully positioned data centers, cross-device FL can utilize up to hundreds of millions of end-user devices spread throughout the globe that use a variety of energy sources. According to a recent study, training a model with FL can emit up to 80 kg of carbon dioxide equivalent (CO2e), which is more than developing a larger transformer model in a centralized training setup using AI accelerators \cite{Wu2022carbon}. Several reasons contribute to the relative inefficiency: the overhead of training on a large-scale, highly heterogeneous collection of client hardware, increased communication costs, and frequently slower convergence \cite{Yousefpour2023}. 
\par
Quantifying and lowering the carbon emissions associated with machine learning (ML) training and inference in a data center environment has gained attention recently \cite{Strubell_Ganesh_McCallum_2020,lacoste2019quantifying, Henderson2020carbon}. Federated Learning's (FL) carbon footprint and the elements that make FL carbon efficient have not yet been fully investigated. Previous studies have provided preliminary results, but they have only provided a partial picture by quantifying the carbon effects of FL in a simulation context or under numerous simplifying assumptions \cite{Qiu2021carbon, Wu2022carbon}. These works refrained from exploring dimensions of the design space toward attaining Green FL, instead concentrating on measures and opportunity sizing. In \cite{Yousefpour2023}, the authors analyzed that a strong correlation exists among all factors and artifacts between an FL task's carbon footprint and the sum of its operating time and the number of users engaged in training or concurrency.
Moreover, the authors concluded that the number of participating clients (concurrency) and training time to achieve the desired accuracy are directly related to the carbon emission of any FL framework. On the other hand, other factors like learning rates, batch sizes, aggregation targets, and local epochs affect how quickly the FL model converges to the target accuracy, affecting how long it takes to finish the training process. Although these factors have no direct impact on carbon emissions, they do indirectly affect the total training speed.
Therefore, these parameters should be included in the multidimensional design of Green FL's ``time'' and ``performance'' aspects. Conversely, concurrency directly impacts time and carbon emissions and hence needs to be taken into account more while designing FL systems to lower carbon emissions. 
\par
The concept of Green Federated Learning was also studied by Yousefpour et al. \cite{Yousefpour2023}. In this study, the authors introduced the notion of Green FL, which entails FL parameter optimization and design decisions to reduce carbon emissions while maintaining competitive performance and training efficiency. Two contributions were made to this work. First, a data-driven methodology to directly measure FL jobs that are executed on a scale in the real world on millions of phones to quantify the carbon emissions of FL. Second, a discussion on difficulties, recommendations, and takeaways from researching the trade-off between energy efficiency, performance, and training time in an FL system. Their findings lay a solid platform for future research in the field of green AI and provide useful insights into how FL can reduce its carbon impact.

\subsection*{Insight on assessing Energy and Carbon Footptint}
The SOTA surveys related to FL frameworks are divided into various challenges: privacy, heterogeneity, client selection, model compression, and communications rounds. Standard accuracy metrics, such as the handling of heterogeneous data, reliability, response time, convergence rate, number of communication rounds, and accuracy of a given model, usually determine SOTA results. However, Energy efficiency or emissions of carbon footprint are frequently disregarded. While the explored metrics are still important, we want to inspire academics to pay attention to other indicators that correspond with the growing concern of the civil community about global warming. Hence, we survey the works where the inclusion of released CO2 is considered a critical parameter in calculating carbon emissions for FL by quantifying it.
 Qiu \etal \cite{Qiu2021carbon} presented the first comprehensive analysis of FL's carbon footprint. The authors proposed a strict model to measure the carbon footprint, making it easier to look into the connection between FL design and carbon emissions. In this work, the authors considered two main procedures to calculate the environmental cost of training deep learning models on edge or in data centers. First, the authors analyzed the energy used by the approach, primarily considering the overall energy used by the hardware. The latter amount is then translated, using geographic coordinates, into CO2 emissions. On the other hand, Yang \etal \cite{Yang2021energy} examined the issue of FL across wireless communication networks: energy-efficient transmission and computing resource allocation. The model under consideration involves the utilization of restricted local computational resources by individual users to train a local FL model using the data they have collected. Subsequently, the trained FL model is transmitted to a federated server (FS), which combines the local FL model and disseminates it to all users. The learning accuracy level affects both computation and communication latencies in FL since it includes users and the BS exchanging a learning model. Meanwhile, both local computing energy and transmission energy must be taken into account during the FL process due to the wireless consumers' limited energy budget. Based on this concept, Qiu \etal \cite{Qiu2023AFL} studied the energy needed by the procedure, primarily considering the overall energy used by the hardware. It consists of communication and training energy for FL and centralized learning. The latter figure is then translated into CO2 emissions according to geographic areas, which will differ significantly depending on the energy sources. Since information on emissions associated with hardware manufacturing is still primarily unavailable, it is not included in this study. Savazzi \etal \cite{Savazzi2023} provided a revolutionary framework for FL that allows for the analysis of energy and carbon footprints. The methodology under consideration measures the energy footprints and carbon equivalent emissions of both fully decentralized, consensus-based systems and vanilla FL solutions. The authors evaluated the estimated carbon and energy footprints while taking into account the effects of different communication environments and the size of the learner community. The suggested framework sheds light on how the various elements of the FL designs affect carbon emissions and energy consumption in terms of system accuracy and number of rounds.

\begin{table}
    \setlength\tabcolsep{2pt} 
\footnotesize  
\centering
    \caption{Green FL-based Applications}
    \label{tab:energyawaeFL} 
    \begin{adjustbox}{width=1\textwidth}
   \begin{tabular}{|l|p{4.5cm}|l|l|l|p{3.5cm}|}
   \hline
    \multirow{2}{*}{Reference} & \multirow{2}{*}{Proposal} & \multicolumn{3}{l|}{Evaluation Criteria} & \multirow{2}{*}{Results} \\\cline{3-5}
    
   {}   & {} & Communication Rounds  & Client Selection  & Model Compression  & {}\\
   \hline
FedGreen \cite{Li2021} & Create a fine-grained gradient compression technique by fusing server-side element-wise aggregation with device-side gradient reduction
& X & X & \checkmark & Give a trade-off analysis between learning accuracy and energy efficiency for FL using gradient compression. \\
GREED \cite{Rana2022} & Propose a client selection mechanism to reduce the energy consumption for IoT devices & X & \checkmark & X & Ascertain that every client who is chosen has enough energy to upload their local models before the deadline by finding the best balance between increasing the number of clients that are chosen and reducing the energy that is taken from their batteries.\\
\cite{Thakur2023energyware} & Propose a client selection algorithm for activity recognition with smartphone &  X & \checkmark & X & chooses the greatest number of clients in order to improve FL's model correctness and convergence rate.\\
\cite{Salh2023smartindustries} & create a productive way to integrate joint edge intelligence nodes in order to reduce energy usage.
&\checkmark & X & X & By jointly optimizing the CPU frequency, improve calculation and transmission to meet FL time and minimize energy usage for IoT devices.
\\
    \hline
    \end{tabular}
\end{adjustbox}
\end{table}

 \subsection*{Application of Green Federated Learning}
Little evidence of energy-efficient FL (green FL) for IoT applications is available in the literature. A few recent studies have proposed integrating FL into IoT systems to provide a sustainable environment, as summarized in Table \ref{tab:energyawaeFL} as follows:
Li \etal \cite{Li2021} suggested FedGreen, which adds fine-grained gradient compression to the original FL to improve it and control the devices' overall energy consumption. The authors introduced the necessary operations to make gradient compression in FL easier, such as server-side element-wise aggregation and device-side gradient reduction. The compressed local gradients' contributions to various compression ratios were assessed using data from a publicly available dataset. Additionally, the authors looked at a trade-off between learning precision and energy efficiency and derived the ideal compression ratio and processing frequency for each device. In order to maximize the number of selected clients and minimize the energy taken from batteries for the selected clients, Rana \etal \cite{Rana2022} proposed a novel client selection called ``EnerGy-AwaRe CliEnt SElection for Green FeDerated Learning (GREED)'' that ensures that all selected clients have enough energy to upload their local models before the deadline. Thakur \etal \cite{Thakur2023energyware} proposed an energy-aware FL framework for activity recognition using a simple client selection algorithm based on the idea that participation of more clients reduces the number of communications rounds. Hence, it reduces energy consumption. Salh \etal \cite{Salh2023smartindustries} proposed an energy-efficient FL framework for green IoT to minimize energy consumption and meet the FL time requirement for all IoT devices. It does this by looking into energy-efficient bandwidth allocation, computing ``Central Processing Unit'' (CPU) frequency, optimization transmission power, and the desired level of learning accuracy. By resolving the bandwidth optimization problem in closed form, the approach effectively optimized the compute frequency allocation and decreased energy consumption in IoT devices. An ``Alternative Direction Algorithm'' (ADA) could resolve the remaining computational frequency allocation, transmission power allocation, and loss to lower energy consumption and complexity at each FL time iteration from IoT devices to edge intelligence nodes.

\section*{FL based Green- Aware IoT Applications}
\label{sec:application}
\begin{table}
    \setlength\tabcolsep{2pt} 
\footnotesize  
\centering
    \caption{FL-based Smart Industry Application}
    \label{tab:FL_industry} 
   \begin{tabular}{|l|p{4.5cm}|l|l|l|p{3.5cm}|}
   \hline
    \multirow{2}{*}{Reference} & \multirow{2}{*}{Proposal} & \multicolumn{3}{l|}{Evaluation Criteria} & \multirow{2}{*}{Results} \\\cline{3-5}
    
   {}   & {} & Security  & Accuracy  & Response Time  & {}\\
   \hline
    \cite{Wang2022smartindustry} & Presented a federated transfer learning framework for cross-domain in smart manufacturing & X & \checkmark & \checkmark & Boost accuracy and response speed in comparison to more complex techniques\\
    \cite{Yazdinejad2022smartindustries} & Create a threat hunting system called block hunter using federated learning to automatically search for attacks in IIoT networks based on blockchain technology for smart factories to prevent cyberattacks. & \checkmark & \checkmark & \checkmark & Improve the accuracy of detection and the speed of response. \\
    \cite{Huong2022smartindustries} & Proposed an interpretable FedX architecture to identify unknown attacks in smart IoT-based factories & \checkmark & \checkmark & \checkmark & Enhance accuracy with faster training time \& less memory consumption\\
    \cite{Fan2023smartindustries} & provide a privacy-preserving data aggregation approach for the IIoT based on federated learning (FLPDA). Data aggregation to defend against reverse analysis threats from industry administration centers and preserve individual user model modifications in federated learning. 
 & \checkmark & X & \checkmark & Compared to current systems, it has less computational, storage, and communication overhead with enhanced security \& privacy. \\
    \cite{Savazzi2023} & Create an analytical framework for energy and carbon footprints for environmentally friendly and sustainable FL-based IIoT systems.  &\checkmark  & \checkmark & \checkmark & A trade-off between energy and test accuracy should be made in light of the model and data footprints for the intended industrial applications.\\
    \cite{Peres2023smartindustries} & Propose conceptual FL framework for collaborative industrial AI applications & \checkmark & \checkmark & \checkmark & Improve security, privacy, and execution speed while lowering computer costs.\\
    \cite{Salh2023smartindustries} & In order to minimize energy consumption and meet the FL time requirement for all IoT devices, develop an efficient integration of joint edge intelligence nodes. It does this by looking into energy-efficient bandwidth allocation, computing Central Processing Unit (CPU) frequency, optimization transmission power, and the desired level of learning accuracy. 
&\checkmark  & \checkmark & \checkmark & By resolving the bandwidth optimization problem in closed form, the approach effectively optimized the compute frequency allocation and decreased energy consumption in IoT devices. An Alternative Direction Algorithm (ADA) could be used to resolve the remaining computational frequency allocation, transmission power allocation, and loss in order to lower energy consumption and complexity at each FL time iteration from IoT devices to edge intelligence nodes.
\\
    \hline
    \end{tabular}
\end{table}

Massive amounts of data are constantly being generated by the growing number of Internet of Things (IoT) devices, but the current cloud-centric approach to IoT big data analysis has sparked worries among the public about data privacy and network costs \cite{Xu2021compression}. In current deployments, a cloud server centrally hosts and processes the real-time data that IoT devices collect. However, the sharing of privacy-sensitive data via cloud-based model training raises concerns from the public and may result in unacceptably high traffic loads and latency \cite{Lim2020application}.
It is better to separate the requirement for remotely gathering and centrally processing the raw data from the model training process in order to lessen these difficulties.
Given these drawbacks of the cloud-centric solution, scientists have recently introduced Federated Learning (FL), a novel method for training a shared global model on datasets decentralized at a loose federation of participating devices (clients) \cite{McMahan2016}. 
\par
However, the training overhead presents FL with a significant hurdle. Popular DNN models, on the one hand, typically have tens of thousands to hundreds of millions of parameters \cite{Sze2017}. The training of the model is highly computationally and memory-intensive due to the constantly growing network complexity and training data \cite{Lim2020application} resulting in a large carbon footprint. However, frequent and high-dimensional parameter changes come with a high cost of communication and can lead to a training bottleneck \cite{Zeyi2018}. perhaps worse, because IoT devices typically have limited computational power, meomory and transmission capacity, the training period will be excessively lengthy or perhaps intolerable. This makes it difficult to quickly implement and use AI models in real-world settings. In this section, we survey the application of green-aware AI based on federated learning for different IoT applications such as smart industry, smart cities, smart healthcare, UAVs, and intelligent transportation.
\begin{table}
    \setlength\tabcolsep{2pt} 
\footnotesize  
\centering
    \caption{FL-based Smart City Applications}
    \label{tab:FL_cities} 
   \begin{tabular}{|l|p{4.5cm}|l|l|l|p{3.5cm}|}
   \hline
    \multirow{2}{*}{Reference} & \multirow{2}{*}{Proposal} & \multicolumn{3}{l|}{Evaluation Criteria} & \multirow{2}{*}{Results} \\\cline{3-5}
    
   {}   & {} & Security  & Accuracy  & Response Time  & {}\\
   \hline
\cite{Otoum2022cities} & In order to achieve network security and trustworthiness in smart cities, present an adaptive framework that combines blockchain technology with federated learning. &
\checkmark & \checkmark & X & Demonstrate the efficacy of the framework in terms of network longevity, energy consumption, and trust using different parameters. The suggested model continues to have high accuracy and detection rates. \\
\cite{Zhang2022cities} & Propose an FL framework to overcome communication overhead issues in the application with smart cities. & X & \checkmark & \checkmark & Demonstrate exceptional training efficiency, precise prediction accuracy, and resilience to erratic network settings.\\
\cite{Algarni2023cities} & Provide a FL-based privacy-aware plan for private remote sensing information.
 & \checkmark & \checkmark & \checkmark & Increase detection rate, accuracy, and convergence speed over current solutions. \\
\cite{Houda2023cities} & Propose DETECT, a promising edge-based architecture that uses FL to secure Internet of Things applications. By protecting the privacy of MEC collaborators and, by extension, the privacy of IoT devices, DETECT enables several MEC domains to work together to jointly and securely reduce IoT assaults.
& \checkmark & \checkmark & X & Utilizing the Edge-IIoTset and NSL-KDD datasets, two well-known IoT attack scenarios, DETECT demonstrates a noteworthy accuracy in terms of both F1 score (87\% in NSL-KDD and 99\% in Edge-IIoTset) and Accuracy (86\% in NSL-KDD and 99\% in Edge-IIoTset).\\
    \hline
    \end{tabular}
\end{table}
\subsection*{Smart Industry}
The Industrial Internet of Things (IIoT) has been widely prevalent due to the rapid development of industrial informatization and the advancements in communication and smart device technology. IIoT technologies make it possible to connect dispersed machines and devices over communication networks, which increases efficiency and productivity. Federated learning (FL) has emerged as a viable approach for privacy-protected, cost-effective, intelligent, green IIoT applications to provide intelligent IIoT services in industries. By aggregating local updates from numerous learning clients, including IIoT devices, FL makes it possible to train high-quality machine learning (ML) models without requiring direct access to the local data. This lessens the chance of privacy leaks. Additionally, FL draws substantial computational and dataset resources from several IIoT devices for ML model training, thereby considerably enhancing the quality of IIoT data training, including accuracy. With limited processing power and data, centralized AI approaches could not accomplish this strategy.  Hence, As indicated in Table \ref{tab:FL_industry}, FL-based IIoT systems in the application with Smart industries are suggested to enhance the functionality, precision, and responsiveness of real-time IIoT applications and produce certain favorable outcomes. 

\begin{table}
    \setlength\tabcolsep{2pt} 
\footnotesize  
\centering
    \caption{FL-based Smart Healthcare Applications}
    \label{tab:FL_health} 
   \begin{tabular}{|l|p{4.5cm}|l|l|l|p{3.5cm}|}
   \hline
    \multirow{2}{*}{Reference} & \multirow{2}{*}{Proposal} & \multicolumn{3}{l|}{Evaluation Criteria} & \multirow{2}{*}{Results} \\\cline{3-5}
    
   {}   & {} & Security  & Accuracy  & Response Time  & {}\\
   \hline
   \cite{Akter2022health}& Create a secure and private FL-based IoMT system. & \checkmark & \checkmark & X & After specific iterations, the findings showed a 90\% accuracy performance, superior to other baseline approaches with accuracy levels of about 80\% using the same epsilon number of 4. Furthermore, this architecture satisfies the privacy preservation paradigm more successfully and quickly. \\
\cite{Qayyum2022health} & Employed the recently developed idea of clustered FL to perform an automated diagnostic of COVID-19. & X & \checkmark & \checkmark & Increase forecast accuracy while cutting down on time expenditure.\\
\cite{Guo2022health} & Provide an FL-based real-time IoMT system for medical data processing. & X & X & \checkmark & Faster service response times for IoMT systems operating in real-time. \\

\cite{Wu2022health} & Provide an innovative cloud-edge FL architecture for in-home health monitoring that protects user privacy by storing user data locally and learning a shared global model in the cloud from numerous houses at the network edges. 
& \checkmark & \checkmark & \checkmark & Derive precise and individualized health tracking by fine-tuning the model using a dataset that is generated from the user's personal, class-balanced data with reduced communication cost. \\
\cite{Zhang2023health} & Provide an end-to-end federated representation learning approach for human mobility predictions To tackle the constraints of heterogeneity in mobility patterns and data scarcity.  & \checkmark & \checkmark & \checkmark & Advantageous over SOTA methods in terms of security, accuracy and response time.\\
\cite{Gupta2023health} & Develop a federated edge-based framework for IoMT devices with limited resources. & X & \checkmark & \checkmark & Reduce training time of videos by 3.6 hrs when employing the split learning phase. Also, achieve 90.32\% global accuracy at the server.\\
\cite{Rehman2023health} & Develop a framework for attack detection based on FL.  & \checkmark & X & X & Increase the attack detection rate accuracy to 80.09\%. \\
\cite{Kalapaaking2023health} & Provide a blockchain-based federated learning solution that protects healthcare systems from poisoning threats by utilizing SMPC model verification.  & \checkmark & X & X & Boost security and privacy for intelligent IoMT applications. \\
\cite{Wang2023health} & Suggest a Privacy Protection Framework for Edge Computing Federated Learning (PPFLEC) IoMT applications & \checkmark & X & \checkmark & Reduce time expenditure by 40\% in comparison to current methods with security measure. \\
\cite{Yu2023health} & In order to overcome the idea drift and convergence instability difficulties in the online federated learning process to recognize physical activities, an unsupervised gradient aggregation technique is created along with an algorithm to calculate unsupervised gradients under the consistency training proposition.
& X & \checkmark & \checkmark & Achieve an average additional 10\% improvement in accuracy and response time.\\
    \hline
    \end{tabular}
\end{table}
\subsection*{Smart Cities} 
The installation of 5G base stations, intercity high-speed rail and urban rail transit, artificial intelligence, and industrial Internet have all an impact on energy conservation and the reduction of CO2 emissions in smart cities \cite{Guo2022smartcity}. Smart cities are actually the result of combining various infrastructure, technologies, and solutions to create intelligent metropolitan areas. Real-time smart IoT systems are one of the main aspects of smart cities. Combining FL and IoT applications is suggested as a means of achieving this, and Table \ref{tab:FL_cities} shows some encouraging outcomes.

\begin{table}
    \setlength\tabcolsep{2pt} 
\footnotesize  
\centering
    \caption{FL-based Unmanned Aerial Vehicles (UAVs) Applications}
    \label{tab:FL_uav} 
   \begin{tabular}{|l|p{4.5cm}|l|l|l|p{3.5cm}|}
   \hline
    \multirow{2}{*}{Reference} & \multirow{2}{*}{Proposal} & \multicolumn{3}{l|}{Evaluation Criteria} & \multirow{2}{*}{Results} \\\cline{3-5}
    
   {}   & {} & Security  & Accuracy  & Response Time  & {}\\
 \hline
\cite{He2023uav} & Provide a FL-based UAV clustering plan & X & \checkmark & \checkmark & Improve clustering stability and shorten service response times\\
\cite{Nasr-Azadani2023uav} & Make a recommendation to support ground wireless networks with FL-based heterogeneous UAVs
& \checkmark & X & \checkmark & Increase response time, network throughput, and learning speed
\\
\cite{Wang2023uav} & Provide a FL-based framework for UAVs to handle resource constraints and heterogeneous data
&X & \checkmark & \checkmark & Improve learning outcomes, reduce computing expenses, and raise UAV trajectory
\\
\cite{Liu2023uav} & By utilizing split learning and FL in a distributed fashion, suggest FL-based UAVs
&X & \checkmark & \checkmark & Boost accuracy rate, convergence, and learning speed
\\
\cite{Yang2023uav} & Provide an FL-based computing system that UAVs can use to track urban transit.
 & X & X & \checkmark & Shorten response times for services, improve network performance, and learning speed.
\\
\cite{Hou2023uav} & Make a security-conscious FL-based learning plan available for UAVs
& \checkmark & X & \checkmark & Boost security while ensuring time and energy savings
\\
\cite{Fu2023uav} & Provide the FL convergence analysis without making any convexity assumptions and show how device scheduling affects the global gradients for UAV application
& X & \checkmark & \checkmark  & In comparison to current benchmarks, the suggested design considerably improves the tradeoff between completion time and test accuracy.
\\
    \hline
    \end{tabular}
\end{table}

\subsection*{Smart Healthcare}
As a new distributed collaborative AI paradigm, FL mainly appeals to smart healthcare since it allows several customers, like hospitals, to coordinate AI training without exchanging raw data. The latest FL ideas for smart healthcare encompass resource-conscious FL, secure and privacy-conscious FL, incentive-focused FL, and personalized FL \cite{Nguyen2022health}. Table \ref{tab:FL_health} discusses several emerging applications of FL in critical healthcare areas, such as COVID-19 detection, medical imaging, remote health monitoring, human activity recognition, and health data management.

\begin{table}
    \setlength\tabcolsep{2pt} 
\footnotesize  
\centering
    \caption{FL-based Smart Transportation Applications}
    \label{tab:FL_transport} 
   \begin{tabular}{|l|p{4.5cm}|l|l|l|p{3.5cm}|}
   \hline
    \multirow{2}{*}{Reference} & \multirow{2}{*}{Proposal} & \multicolumn{3}{l|}{Evaluation Criteria} & \multirow{2}{*}{Results} \\\cline{3-5}
    
   {}   & {} & Security  & Accuracy  & Response Time  & {}\\
   \hline
\cite{Kang2021transport} & Provide an FL and Blockchain-based work distribution plan for smart transportation systems & \checkmark & X & X & The suggested methods have the potential to significantly increase federated edge learning security \\
\cite{Sepasgozar2022transport} & Provide an FL method for Network Traffic Prediction (Fed-NTP) that trains the model locally using the Long-Short-Term Memory (LSTM) algorithm to accurately predict network traffic flow while maintaining privacy & \checkmark & \checkmark & X &Boost network traffic flow accuracy forecast rate, privacy, and security \\
\cite{Liu2023transport}& Create a sophisticated multi-tier FL-based vehicle traffic forecasting system for intelligent transportation networks
& X &\checkmark & \checkmark & Boost the prediction accuracy rate and system efficiency \\
\cite{Xu2023transport}& provides an asynchronous federated learning technique with a dynamic scaling factor that is based on blockchain
& \checkmark& \checkmark & \checkmark & Verify superior dependability, efficiency, and learning performance\\
\cite{Yuan2023transport}& Provide a FL-IoT solution for smart city traffic flow prediction & X & \checkmark & \checkmark & Increase forecast accuracy while cutting down on time expenditure\\
\cite{Hijji2023transport}& provide an intelligent hierarchical framework for maintaining road infrastructure   & \checkmark & \checkmark & \checkmark & Support large-scale real-time smart transportation systems with adaptive resource allocation
\\
    \hline
    \end{tabular}
\end{table}

\subsection*{Unmanned Aerial Vehicles (UAVs)}
Federated learning (FL), in conjunction with unmanned aerial vehicles (UAVs), has been viewed as a promising paradigm for handling the enormous volumes of data produced by IoT devices. Table \ref{tab:FL_uav} represents some recent FL-based UAV applications.

\subsection*{Smart Transportation}
The Internet of Things (IoT), sensor technologies, and communication technologies have developed rapidly, spurring the growth of smart transportation systems (STS). However, because vehicle networks are dynamic, it is difficult to decide on vehicle behaviors in a timely and correct manner. Furthermore, there is always a risk to the security and privacy of vehicle information when mobile wireless connections are present. 
One possible method to improve the effectiveness and safety of intelligent transportation systems is federated learning \cite{Zhang2023transport}. Although FL has been examined in great detail, more is needed to understand the communication and networking issues that arise when FL is used in densely populated but dynamic vehicle networks. The limited storage and communication capacities of individual vehicles hinder the timely training of an FL model in distributed vehicular networks \cite{Sangdeh2023transport}. Here, we examine a few common FL-based ITS systems, as shown in Table \ref{tab:FL_transport}.

\section*{Discussions \& Future Scope}
\label{sec:futurescope}
Various surveys in the literature established the requirement of FL frameworks in IoT applications for security and privacy preservation. According to our findings, other challenges are associated with any IoT application, as IoT or edge devices are resource-constrained in terms of memory, computational power, and energy. 
In this survey, we initially establish the critical challenges related to FL frameworks for IoT applications and discuss several existing solutions for them. However, only some existing solutions address the challenges related to green-aware AI for a sustainable environment. Then, we establish that energy and carbon footprint analysis is essential for FL-based IoT applications to ensure a sustainable environment. Some previous works studied the energy and carbon footprint in FL frameworks, but they still needed to study it for IoT applications. Few recent studies have focused on energy consumption using reducing communication rounds, client selection algorithms, or model compression. However, carbon emission studies are missing in their proposed FL framework for IoT applications. As we established in our survey, analyzing carbon footprint is essential for a sustainable environment. Move from this consideration, we finally survey some of the FL-based IoT applications to establish the potentiality of green FL frameworks. 

\par
Even if the FL has proliferated in recent years, there is still a need for this study area to be improved, and new approaches and frameworks can be created to maximize energy efficiency and minimize the carbon footprint. Several prospective areas and challenges for future study are mentioned as follows:
\begin{itemize}
    \item A way to effectively address the challenges described in Section \ref{sec:challenges} to FL frameworks for efficiency and effectiveness should be determined. By offering a fundamental foundation for efficient model learning, FL enables users to work together to train a global model with their own datasets. However, there is no guarantee of effectiveness due to statistical and system heterogeneity of data and client devices. To address statistical heterogeneity, data moderation, personalization, and cluster mechanisms have been widely developed in a variety of FL frameworks (Section \ref{subsec:effectiveness}). Nevertheless, the proposed techniques used in the FL frameworks frequently reduce accuracy or effectiveness in addressing statistical and system heterogeneity. Therefore, while addressing heterogeneity in the FL, it is essential to weigh the trade-off between accuracy and energy efficiency.

    \item Numerous methods have been put forth to increase federated learning's communication effectiveness described in Section\ref{sec:variants}. These methods mostly focus on reducing the size of the model updates that are sent or the frequency at which these updates are communicated. Nevertheless, the advantages of using such solutions frequently come at the expense of a reduction in the accuracy of the classification or forecast. To create effective and efficient communication models for federated learning, a thorough examination of the trade-off between accuracy preservation and communication efficiency should be done. More precisely, we require novel communication strategies that can demonstrate efficiency at the Pareto frontier—that is, strategies that can provide an accuracy level greater than any other strategy at the same communication budget and over a wide range of communication/accuracy profiles.

    \item Key issues that impact FL performance include communication and computation efficiency described in Section \ref{sec:challenges}. To address these issues, it is required to reduce the communication rounds, use an efficient client selection algorithm, or compress the deep learning model. Several techniques to reduce the communication rounds, client selection, and model compression are proposed in the literature described in Section \ref{sec:variants}. However, there is a lack of evidence to maintain the tradeoff between accuracy, communication efficiency, and energy efficiency in a heterogeneous environment. Therefore, while addressing effectiveness in the FL, it is essential to weigh the trade-off between accuracy, communication efficiency, and energy efficiency.
    
    \item Energy and carbon footprint analysis is a key ingredient to propose an FL-based IoT application for a sustainable environment described in Section \ref{sec:greenFL}. In the literature, there is no evidence available for energy and carbon footprint analysis for FL-based IoT applications. Therefore, while maintaining effectiveness and efficiency, it is essential to analyze the energy and carbon footprint to validate green-aware AI. Hence, there is a need for numerical assessment of many communication- and computationally-efficient FL techniques for IoT applications in terms of energy use and carbon equivalent emissions.

\end{itemize}
\section*{Conclusion}
In this study, we conduct a thorough analysis of FL driving for emerging IoT applications to create environmentally friendly and sustainable industrial facilities in the future. FL is a newly developed AI training model that is garnering the greatest attention for achieving IoT application capabilities like scalability and security and privacy concerns. However, the impact of energy efficiency and carbon emission for FL-based IoT applications is not discussed in the previous surveys.
The incomplete understanding of the application of green FL in IoT domains is the driving force for this investigation. To address this, we first provide specific challenges and solutions for fusing IoT systems and the FL method, highlighting recent advancements in both areas. Then, we offer an extensive overview of the most recent developments of standard FL frameworks that can be used for IoT systems in a number of critical domains, including smart cities, intelligent transportation, smart medical, UAVs, and industrial IoT systems. Several key lessons have been examined and compiled from the survey data. Ultimately, this paper highlights the primary obstacles and potential avenues for future investigation. In addition to encouraging research activities targeted at lowering the energy consumption and carbon footprint of IoT devices and implementing real-time smart IoT applications, our work is intended to draw more attention to this promising subject.

\bibliographystyle{ACM-Reference-Format}
\bibliography{bibfile}

\end{document}